\documentclass[runningheads]{llncs}
\usepackage{graphicx}
\usepackage{subfigure}
\usepackage{amsmath}
\usepackage{array}
\usepackage{booktabs}
\usepackage{multirow}
\usepackage{cite}
\usepackage{float}
\usepackage[misc]{ifsym} 

\begin{document}
\title{ 
 Hierarchical Interaction Networks with 
 Rethinking Mechanism
for Document-level Sentiment Analysis}
\titlerunning{Hierarchical Interaction Networks with Rethinking Mechanism for DSA}


\author{Lingwei Wei\inst{*1,3} \and
Dou Hu\inst{*2}\Letter \and
Wei Zhou\inst{1}\Letter \and
Xuehai Tang\inst{1} \and
Xiaodan Zhang\inst{1} \and 
Xin Wang\inst{1} \and 
Jizhong Han\inst{1} \and 
Songlin Hu\inst{1,3}}

\authorrunning{L. Wei et al.}

\institute{Institute of Information Engineering, Chinese Academy of Sciences
 \and
 National Computer System Engineering Research Institute of China
\\
\and
School of Cyber Security, University of Chinese Academy of Sciences
\\
\email{hudou18@mails.ucas.edu.cn}, \email{zhouwei@iie.ac.cn}
}

\maketitle

\begin{abstract}
    \let\thefootnote\relax\footnotetext{*  Equal Contribution.}
    \footnotetext{\Letter {} Corresponding author.}
    
    Document-level Sentiment Analysis (DSA) is more challenging due to vague semantic links and complicate sentiment information. 
    Recent works have been devoted to leveraging text summarization and have achieved promising results. 
    However, these summarization-based methods did not take full advantage of the summary including ignoring the inherent interactions between the summary and document.
    As a result, they limited the representation to express major points in the document, which is highly indicative of the key sentiment.  
    In this paper, we study how to effectively generate a discriminative representation with explicit subject patterns and sentiment contexts for DSA. 
    A Hierarchical Interaction Networks (HIN) is proposed to explore bidirectional interactions between the summary and document at multiple granularities and learn subject-oriented document representations for sentiment classification. 
    Furthermore, we design a Sentiment-based Rethinking mechanism (SR) by refining the HIN with sentiment label information 
    to learn a more sentiment-aware document representation. 
    We extensively evaluate our proposed models on three public datasets. The experimental results consistently demonstrate the effectiveness of our proposed models and show that HIN-SR outperforms various state-of-the-art methods.

\end{abstract}

\section{Introduction}
Document-level Sentiment Analysis (DSA), a subtask of Sentiment Analysis (SA), aims to understand user attitudes and identify sentiment polarity expressed at the document level.
This task has grown to be one of the most active research areas in natural language processing and plays an important role in many real-world applications such as intent identification \cite{wang2016influence}, recommender systems\cite{hyun2018review} and misinformation detection\cite{rumor_yuan_2019, yuan2019learning}.

Generally, DSA can be regarded as a text classification problem and thus traditional text classification approaches \cite{kim-2014-convolutional, tang-etal-2016-effective, devlin-etal-2019-bert, DBLP:conf/cncl/SunQXH19} can be adopted naturally.
Different from other subtasks in SA, 
DSA is more challenging due to the large size of words,
vague semantic links between sentences and abundance of sentiments.
Hence, the research question that how to 
learn an expressive document representation to understand long documents for sentiment classification has been given growing interest to researchers. 

Inspired by the document structure, 
one of the earliest and most influential models
HAN was proposed by \cite{yang2016hierarchical} 
to encode the entire document, which suffered from 
attending explicit but irrelevant sentimental words.
Subsequent works mainly dedicated to 
introducing latent topics \cite{DBLP:journals/ipm/PergolaGH19}
or global context \cite{DBLP:journals/corr/abs-1908-06006}
to tackle this limitation.
However, most of the user-generated documents are very comprehensive and contain a wealth of sentiments,
which makes it difficult to directly learn from the whole document without the understanding of the major points,
especially in long documents. 
The above works attempted to explore the major points of the document by learning a global embedding from the document.
Intuitively, the user-generated summary contains more accurate information about the major points of the document.
These summaries describe the long document in a more specific way,
which are highly indicative of the key sentiment and subject, and can facilitate further to identify important 
text present and sentiments.

To reduce the processing of the substantial text in the document and 
be well aware of the major idea of it, summary-based methods\cite{hole2013real, mane2015summarization} introducing the user-generated summary has been developed for DSA and achieved promising results, which brings brilliant processing for understanding complex documents.
They refined the document with an abstractive summary to predict the sentiment effectively.
Recently, some effective works \cite{DBLP:conf/ijcai/MaSLR18, pmlr-v95-wang18b} concerned both the text summarization task and DSA, and jointly modeled them to boost each other.
Nevertheless, most of the joint models did not fully utilize user-generated summaries and ignored the interaction between summary and document, because they did not encode summaries explicitly during test time.

For example, the document of a product review in Amazon SNAP Review Dataset is ``...They just sent a new camera and it showed up without any warning or communication about the \underline{bad} one. \underline{Minimal} \textit{Customer Service}...The 1st camera was \underline{promising} and worked so \underline{well} for about two weeks..." and its length is $1134$. The corresponding summary is ``Quality is a reflection of \textit{Customer Service}". 
The document contains complex sentiment-expressed, such as {\it promising} or {\it bad} corresponding to positive or negative, respectively.
The subject {\it Customer Service} can help better predict the key sentiment of the document, 
such as {\it bad} and {\it minimal}. 
Meanwhile, the document can supplement ambiguously semantic features in the summary, such as the details of {\it Customer Service}.
They are complementary.
Therefore, the auxiliary of the summary is significant for subject mining and semantic understanding in DSA.

To tackle the aforementioned problems, 
we investigate how to effectively focus on more accurate subject information for DSA. 
In this paper, we present an end-to-end model, named {H}ierarchical {I}nteraction {N}etwork (HIN), to encode the bidirectional interactions between summary and document. 
The method works by utilizing multi-granularity interactions between summary and document, accordingly to learn a subject-oriented document representation. 
Specifically, the interactions at the character level and the contextual semantic features can be captured via BERT\cite{devlin-etal-2019-bert}.
Afterward, the segment-level interactions are encoded by a gated interaction network and the context of the document is taken into account simultaneously.
Finally, the document-level interactions are embedded via the attention mechanism to learn a more expressive document representation with the consideration of subject information for predicting sentiments.

Furthermore, because of the complex sentiment in the document, we attempt to learn the affective representation and alleviate the distraction from other sentiments. 
We introduce the sentiment label information into the model in a feedback way.
Most existing structures learn document representation via only feedforward networks and have no chance to modify the invalid features of the document.
Some effective works in image classification \cite{DBLP:journals/corr/abs-1708-04483} and named entity recognition \cite{DBLP:conf/ijcai/GuiM0ZJH19} added feedback structure to re-weight of feature embeddings and obtained gratifying results. 
Motivated by their works, we propose a {S}entiment-based {R}ethinking mechanism (SR) and feedback on the sentiment polarity label information. 
This rethinking mechanism can refine the weights of document embeddings to learn a more discerning low-level representation with guidance from the high-level sentiment features, and relieve the negative effects of noisy data simultaneously. 

We evaluate our proposed models on three public datasets from news to reviews. Through experiments versus a suite of state-of-the-art baselines, we demonstrate the effectiveness of interactions and the rethinking mechanism, and the model HIN-SR can significantly outperform baseline systems.

The main contributions of this paper are summarized as follows.
\begin{itemize}
    
  \item We present HIN to incorporate 
  multi-granularities interactions between summary and document-related candidates and learn subject-oriented document representations for document-level sentiment analysis. To the best of our knowledge, we are the first neural approach to DSA that models bidirectional interactions to learn the document representation.

  \item We propose SR to adaptively refine the HIN with high-level features 
  to learn a more discriminative document representation and can relieve the negative impacts of noisy samples simultaneously.

  \item We evaluate our model on three public datasets.  
  Experimental results demonstrate the significant improvement of our proposed models over state-of-the-art methods.
  The source code and dataset are available\footnote{https://github.com/zerohd4869/HIN-SR}.

\end{itemize}

The remainder of this paper is structured as follows. We review the related works in Section \ref{t:related}. Then we explain the details of our contributions in section \ref{t:model}. Section \ref{t:exp} describes the experiments and the results. Further analysis and discussion are shown in Section \ref{t:analysis}. In the end, the conclusions are drawn in Section \ref{t:con}.

\section{Related Works} \label{t:related}
\subsection{Document-level Sentiment Analysis}

In the literature, 
traditional methods regarded document-level sentiment analysis (DSA) as a text classification problem. 
Therefore, typical text classification approaches have been
naturally applied to solve the DSA task, such as 
Bi-LSTM \cite{tang-etal-2016-effective}, TextCNN \cite{kim-2014-convolutional}
and BERT \cite{devlin-etal-2019-bert}.

To learn a better document representation, 
some effective methods have been proposed.
Tang et al.\cite{tang-etal-2015-document} encoded the intrinsic relations between sentences in the semantic meaning of a document. 
Xu et al.\cite{DBLP:conf/emnlp/XuCQH16} proposed a cached LSTM model to capture the overall semantic information in a long text.
Yang et al.\cite{yang2016hierarchical} firstly proposed a hierarchical attention network (HAN) for document classification, which can aggregate differentially informative components of a document.
Remy et al.\cite{DBLP:journals/corr/abs-1908-06006} extended HAN to take content into account.
Song et al.\cite{DBLP:conf/icde/Song19} integrated user characteristics, product and review information to the neural networks.
Although most of the methods attempted to explore the major idea of the document consisting of comprehensive information via a learnable embedding, they obtained the major points from a user-generated summary directly would be more accurate than learning from the entire document.

Some effects based on text summarization have been devoted to predicting the sentiment label. Text summarization task naturally compresses texts for an abstractive summarization from the long document.
Rush et al.\cite{DBLP:conf/emnlp/RushCW15} first proposed an abstractive-based summarization model, adopting an attentive CNN encoder for distillation.
{Bhargava} and {Sharma}\cite{bhargava2017msats} leveraged different techniques of machine learning to perform sentiment analysis of different languages.
Recently, there are some studies concerning both text summarization and document-level sentiment classification. 
Ma et al.\cite{DBLP:conf/ijcai/MaSLR18} proposed a hierarchical end-to-end model HSSC for joint learning of text summarization and sentiment analysis, exploiting the sentiment label as the further summarization of text summarization output.
Based on their work, Wang et al.\cite{pmlr-v95-wang18b} extended a variant hierarchical framework with self-attention to capture emotional information at the text and summary levels.
Although these methods boosted performance by jointly encoding the summary and document, they neglected the bidirectional interactions between the user-generated summary and document.

In this paper, we attempt to capture interactions between the summary and document from multiple granularities to learn a subject-oriented document representation for DSA.

\subsection{Rethinking Mechanism}
Previous works attempting to use a rethinking mechanism have demonstrated promising results in
image classification \cite{DBLP:journals/corr/abs-1708-04483} 
and named entity recognition \cite{DBLP:conf/ijcai/GuiM0ZJH19}.
Li et al.\cite{DBLP:journals/corr/abs-1708-04483} applied the output posterior possibilities of a CNN to refine its intermediate feature maps.
Gui et al.\cite{DBLP:conf/ijcai/GuiM0ZJH19} used the high-level semantic feedback layer to refine the weights of embedded words and tackled the word conflicts problem in the lexicon.

Inspired by their works, we extend these concepts into sentiment analysis and design a rethinking mechanism to introduce the sentiment label information to learn an expressive document representation with more sentiment features and alleviate the negative impacts of the data noise.

\section{Method} \label{t:model}
\begin{figure}[t]
    \centering
    \includegraphics[width=0.9\linewidth]{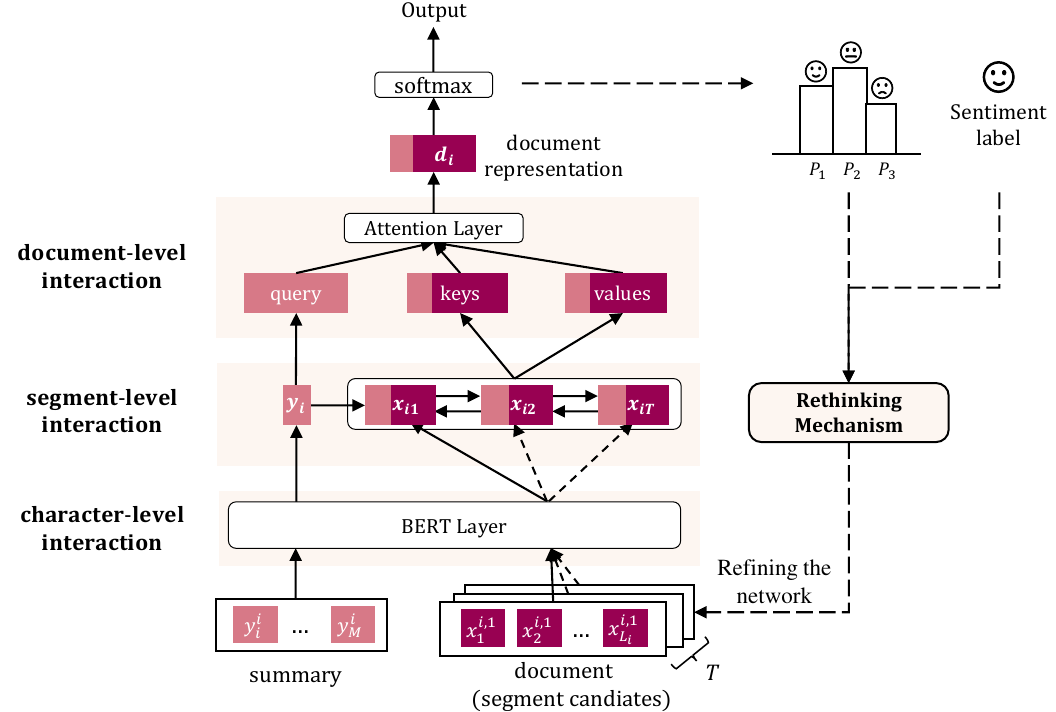}
    \caption{\label{fig2} An illustration of the proposed HIN-SR.
    It consists of a hierarchical interactions network and a sentiment-aware rethinking mechanism.
    In the figure, we show that the document is distilled into $3$ segment candidates with the guidance of the summary.
     }
    \label{fig_framework}
  \end{figure}

In this section, we describe a unified architecture for document-level sentiment analysis.
Firstly, we define some notations in section \ref{sp}.
Then, we introduce the architecture of
 the hierarchical interactions networks (HIN) model in section \ref{hin}. 
 Finally, we show details of the sentiment-based rethinking (SR) mechanism in section \ref{sr}.
 An illustration of the overall architecture in this paper is shown in Fig.~\ref{fig2}.

\subsection{Problem Definition} \label{sp}

Given a dataset that consists of $D$ data samples,
the $i$-th data sample $(x^i,y^i,l^i)$ contains 
a user-generated document $x^i$, the corresponding summary $y^i$, and the corresponding sentiment label $l^i$.
The model is applied to learn the mapping from the source text $x^i$ and $y^i$ to $l^i$.
Before assigning the document to the model, we attempt to reconstruct the document to purify the complex document by leveraging the user-generated summary.  
This preprocess can enable the model to pay more attention to shorter text relevant to the major idea.
We compress the document $x^i$ into several segment candidates 
$x^{i} = \{ x^{i,1},x^{i,2},...,x^{i,T} \}$ 
  with high text similarity given the query (i.e., summary $y^i$),
  where $T$ is the number of candidates.
Specifically, each candidate $x^{i,j}$ and the summary $y^{i}$ are sequences of characters:
$
x^{i,j} = \{x^{i,j}_1,x^{i,j}_2,...,x^{i,j}_{L_i} \}$,
$
y^{i} = \{ y^{i}_1,y^{i}_2,...,y^{i}_{M_i} \},
$
where $L_{i}$ and $M_i$ denote the number of characters in the sequences $x^{i,j}$ and $y^{i}$, respectively.
The label $l^i \in \{ 1,2,...,K \}$ denotes the sentiment polarities of the original document $x^{i}$,
from the lowest rating $1$ to the highest rating $K$.

\subsection{Hierarchical Interactions Networks} \label{hin}
 As shown in Fig.~\ref{fig2}, 
 the proposed model HIN is composed of several components including
 {\it character-, segment-}, and {\it document-level interaction encoding} and {\it decoding}.

\subsubsection{Character-level Interaction Encoding.}

In this part, 
we employ the BERT \cite{devlin-etal-2019-bert} to model character-level interaction between
the summary $y^i$ and segment candidates $x^{i}$ generated by the document.
BERT is a masked language model that 
apply multi-layer bidirectional transformer and large-scale
unsupervised corpora to obtain contextual semantic representations, which has achieved promising performance in many tasks. 
The summary $y^i$ and segment candidates $x^{i}$ are as an input-pair sequence ({\it summary, candidate}) to feed into BERT.

Formally, we concatenate two character-based sequences of the summary $y^i=\{ y^{i}_1,y^{i}_2,...,y^{i}_{M_i} \}$ and $j$-th candidate $x^{i,j} = \{ x^{i,j}_1,x^{i,j}_2,...,x^{i,j}_{L_i} \}$ into a sentence-pair text sequence $\{[CLS], y^i, [SEP], x^{i,j}, [SEP] \}$. $[CLS]$ and $[SEP]$ represent the beginning and ending of the sequence, respectively. Then we pad each text sequence in a mini-batch to $N$ tokens, where $N$ is the maximum length in the batch. 
After finetuning the training parameters,
the outputs of BERT are denoted as the summary vector
$\textbf{y}^c_i$ and the candidates' vectors $\textbf{x}_{ij}, j \in [1,T]$, where $T$ is the number of the candidate.

\subsubsection{Segment-level Interaction Encoding.}

Then, we design the gated interaction network by applying a bidirectional gated recurrent unit(Bi-GRU) to embed the segment-level interactions between the summary and the candidate.
The vector including both summary and candidate features propagates through the hidden units of GRU.
For each document $d_i$, we concatenate the summary vector $\mathbf{y}^c_i$ and $j$th candidate vector $\mathbf{x}_{ij}$ as input to compute the update hidden state by

\begin{equation}
\overrightarrow{\mathbf{h}_{ij}} = \overrightarrow{\text{GRU}}
([\mathbf{y}^c_i;\mathbf{x}_{ij}]),   j \in [1,T],
\end{equation}
\begin{equation}
\overleftarrow{\mathbf{h}_{ij}}= \overleftarrow{\text{GRU}}
([\mathbf{y}^c_i;\mathbf{x}_{ij}]),   j \in [T,1].
\end{equation}

We concatenate $\overrightarrow{\mathbf{h}_{ij}}$ and $\overleftarrow{\mathbf{h}_{ij}}$ to get an annotation of $j$th candidate,
{ \it i.e.}, $\mathbf{h}_{ij}=[\overrightarrow{\mathbf{h}_{ij}} ;\overleftarrow{\mathbf{h}_{ij}}]$, 
which summaries contextual segment-level interactive features among summary and each candidate globally.
Simultaneously, the module aggregates contextual semantic information of segments
and captures structural features of the document.
The embedding of summary $\mathbf{y}^s_i$ at the segment level is defined by an average-pooling operation of the output of BERT:
\begin{equation}
    \mathbf{y}^s_i = \frac{1}{T}  \sum_{j=1}^{T} { 
        tanh(\mathbf{W}_s 
        \mathbf{h}_{ij}+\mathbf{b}_{s}) }.
\end{equation}

\subsubsection{Document-level Interaction Encoding.}
Not all segments contribute equally to the sentiment of the document.
Hence, we extract the embedding of summary from text encoder
and adopt an attention mechanism to formulate the document-level interaction  
by 
extracting subject-oriented segments that are essential to the sentiment:

\begin{equation}
\mathbf{u}_{ij} = tanh(\mathbf{W}_d \mathbf{h}_{ij}+\mathbf{b}_d),
\end{equation}
\begin{equation}
    \alpha_{ij}= \frac {exp(\mathbf{u}_{ij} \mathbf{y}^s_{i} )} {\sum^T_{j=1}{{exp(\mathbf{u}_{ij} \mathbf{y}^s_{i} )}}},
\end{equation}

\begin{equation}
    \mathbf{d}_i=\sum^T_{j=1}{\alpha_{ij}\mathbf{h}_{ij}}. 
    \label{di}
\end{equation}
Specifically, we first feed the candidate annotation $\mathbf{h}_{ij}$ 
through a dense layer 
to get $\mathbf{u}_{ij}$ as a hidden representation of $\mathbf{h}_{ij}$.
We evaluate the contribution of each candidate in the document 
as the similarity with the vector of summary $\mathbf{y}_i$
and formulate a normalized importance weight $\alpha_{ij}$ 
through a {\it softmax} function.
And then, the representations of informative segments are aggregated to form a document vector $\mathbf{d}_i$.

\subsubsection{Decoding and Training.}

Given a document $x$ and its corresponding summary $y$, we obtain final document vector $\mathbf{d}$ with subject and context sentiment information.
We can use a $softmax$ function to project it into sentiment distribution.
The probability distribution of the sentiment label can be computed as:

\begin{equation}
    \hat{P}(l|x, y) = softmax(\mathbf{W}_c \mathbf{d} + \mathbf{b}_c).
\end{equation}
The logistics layer makes the final
prediction with the top probability of the sentiment label.

We employ the cross-entropy error to train the model by minimizing the loss. 
The loss function can be defined as:
\begin{equation}
    L = - \sum_{l=1} ^ K
     {p}({l},x,y) \text{log} \hat{P}(l|x,y),
\end{equation}
where $\hat{P}(l|x,y)$ is the probability distribution of labels,
and ${p}({l},x,y)$ is the gold one-hot distribution of sample $(x,y)$.

\subsection{Sentiment-based Rethinking Mechanism} \label{sr}
Although HIN can effectively learn an expressive document representation,
it still has insufficiency to capture sentiment features for sentiment analysis.
Intuitively, when two samples of different sentiment labels have similar posterior probabilities,
it is not easy to classify them.

In this part, we introduce the gold sentiment label information 
as the high-level features to guide the previous layers 
based on the current posterior probabilities of these confusable categories.
As a result, the bottom layers can be strengthened or weakened 
to capture more discriminative features specifically for those 
categories difficult to distinguish.
When an input sample passes through HIN, instead of immediately making a classification based on 
the predicted posterior probability of the sample belonging to a specific sentiment,
a rethinking mechanism is deployed to propagate the predicted posterior probability to 
the bottom layers to update the network.

Formally, we denote the state vector as $\mathbf{s}$ 
to represent the current document representation information and input it to the rethinking mechanism.
The state of $i$-th document 
is computed by: 
\begin{equation}    
    \mathbf{s}_i = \mathbf{d}_i, 
\end{equation}
where $\mathbf{d}_i$ is the document representation by Eq.(\ref{di}).
A feedback layer is a fully connected layer, that is:
\begin{equation}
   \hat{P}(l|x,y) =  softmax(\mathbf{W}_r \mathbf{s}_i + \mathbf{b}_r),
\end{equation}
 where $\mathbf{W}_r$ and $\mathbf{b}_r$ are the parameters.

Then we introduce the sentiment label information as high-level features to feedback a reward of the current state. 
In the $t$-th episode,
the reward of the sample 
is defined:
\begin{equation}
    r^{(t)} = \lambda  r^{(t-1)} + (1-\lambda) {p}(l, x, y)  \hat{P}({l} | x, y), 
\end{equation}
where $r^{(t-1)}$ represents the reward of the last episode and the reward initialization of training samples is $r^0=1$,  
$\lambda \in [0,1] $ is a hyperparameter to 
weight the importance of the sentiment label of the current $t$-th episode against the previous $t-1$ step.
${p}({l},x,y)$ is the gold one-hot distribution of sample $(x,y)$.
We utilize the reward to reweight the lower-level features in the HIN
to enable it to selectively emphasize some discriminative sentiment features, 
and suppress the feature causing confusion in the classification.

Specifically, we rethink the cross-entropy error between gold sentiment distribution
and the sentiment distribution of HIN with cumulative rewards. 
In $t$-th episode, the loss function is:
\begin{equation}
L' = -
\sum_{l=1} ^ K
r^{(t)}
  {p}({l},x,y) log \hat{P}(l|x,y).
\end{equation}

\section{Experiments} \label{t:exp}

In this section, we first describe the experimental setup. We then report the results of experiments and demonstrate the effectiveness of the proposed modules.

\subsection{Experimental Setup}

\subsubsection{Datasets.}
We conduct experiments on three public benchmark datasets, one from the online news and another two from the online reviews. The details are shown in Table \ref{tab:dataset}.

\begin{itemize}    
    \item 
    \textbf{Toys \& Games} and \textbf{Sports \& Outdoors} are parts of 
    the {S}tanford {N}etwork {A}nalysis {P}roject(SNAP)\footnote{http://snap.stanford.edu/data/web-Amazon.html}, provided by  \cite{he2016ups}. These datasets consist of reviews from Amazon spanning May 1996-July 2014. 
    Raw data includes review content, product, user information, ratings, and summaries. 
    The rating of a product is from 1 to 5.
    Following existing works \cite{DBLP:conf/ijcai/MaSLR18,pmlr-v95-wang18b},  
    we select the first 1,000 samples for validation, 
    the following 1,000 samples for testing, and the rest for training.
    
    \item
    \textbf{Online News Datasets (News)}   
      is from {\it Emotional Analysis of Internet News} task\footnote{
        https://www.datafountain.cn/ competitions/350}. 
    The dataset is collected from websites including news websites, WeChat,
     blogs, Baidu Tieba, etc.
    Each document consists of a title, main-body content and sentiment polarity.
    Here, the title is used as the summary. 
    We randomly select 80\% of data for training, 10\% for validation and the remaining 10\% for testing \cite{yang2016hierarchical}.
\end{itemize}

\begin{table}
  \centering
  \caption{Statistics of three datasets. \# denotes the average length.
  }\label{tab:dataset}
  % \resizebox{\linewidth}{!}{$
  \begin{tabular}{|c|c|c|c|c|}
  \hline
  Dataset            & Total size & Classes & \# Summary & \# Document \\      
  \hline 
  Toys \& Games      & 167,597   & 5       & 4.4        & 99.9     \\
  Sports \& Outdoors & 296,337   & 5       & 4.2        & 87.2 \\
  News               & 14,696    & 3       & 23.8        & 1216.1 \\
  \hline
  \end{tabular}
\end{table}

\subsubsection{Implementation Details.}
In the implementation, we first preprocess to compress the document into 
$3$ segment candidates empirically, i.e., $T=3$. 
We use the bert-based model
  (uncased, 12-layer, 768-hidden, 12-heads, 110M parameters)
 for reviews dataset,
and the bert-base-chinese model 
 (12-layer, 768-hidden, 12-heads, 110M parameters) 
 for news dataset \cite{devlin-etal-2019-bert}.  
We train the model using Adam optimizer with the learning rate of $5e\text{-}6$ for 2 epochs. 
The dropout probability of Bi-GRU is $0.1$. 
We set the hyperparameter $\lambda$ to $\{0.6, 0.8\}$ and maximum length 
$N$ to $256$.

\subsection{Compared Methods}

We compare HIN and HIN-SR with several baselines, including traditional methods and summary-based methods for DSA. 
The details are as follows.

\begin{itemize}
    \item \textbf{TextCNN} \cite{kim-2014-convolutional}. 
    The method used CNN to learn document representations.   
    \item \textbf{Bi-LSTM}  \cite{tang-etal-2016-effective}.
    This was the baseline that directly took the whole document as a single sequence
    using a bidirectional LSTM network for DSA. 
    \item \textbf{BERT} \cite{devlin-etal-2019-bert}. This was a pre-trained model with a deep bidirectional transformer. 
    \item \textbf{BERT(head+tail)} \cite{DBLP:conf/cncl/SunQXH19}. The model finetuned BERT with different truncated methods. 
    \item \textbf{HAN} \cite{yang2016hierarchical}. The method classified documents via hierarchical attention networks.
    \item \textbf{CAHAN} \cite{DBLP:journals/corr/abs-1908-06006}. The method extended the HAN by making sentence encoder context-aware.
    \item \textbf{HSSC} \cite{DBLP:conf/ijcai/MaSLR18}. The method applied a hierarchical end-to-end model for joint learning of text summarization and sentiment classification.
    \item \textbf{SAHSSC} \cite{pmlr-v95-wang18b}. The method jointly established text summarization and sentiment classification via self-attention. 
    
\end{itemize}

\subsection{Experimental Results}

Experimental results are illustrated in Table \ref{tab:result}, divided into three blocks: traditional methods, summary-based methods, and our proposed models.  
Our proposed models perform the best among all baselines on three datasets, which reveals the effectiveness and advancement for DSA.

The first block shows the comparative results with traditional methods.  
HIN and HIN-SR outperform traditional methods, and the benefit mainly comes from the full utilization of the user-generated summary, boosting a more discriminative document representation.
As expected, HAN and CAHAN using Bi-LSTM with hierarchical attention 
achieve better results than traditional classification methods 
TextCNN and Bi-LSTM, 
which confirms the effectiveness of hierarchically modeling documents.
It is observed that BERT and BERT(head+tail) achieve mediocre results,
which reveals that they can capture contextual semantic information effectively.

Then, we compare our models with summary-based methods including HSSC and SAHSSC.
The results in the second block of Table \ref{tab:result}
indicate that HIN-SR outperforms among these summary-based baselines on both Toys \& Games and Sports \& Outdoors datasets.
It is not surprising since HIN  
 takes advantage of the user-generated summary via modeling the interactions between the summary and document
at multiple granularity levels and explores more discriminative features.

Moreover, HIN-SR outperforms HIN among three datasets.
The result demonstrates the positive impacts of the sentiment-based rethinking mechanism,
which can further leverage
the sentiment-aware document representation to refine the weights of document features
and tackle the negative impacts of data noise.

\begin{table}[h]
  \centering
  \caption{Experimental results on three datasets.  
  Evaluation metric is accuracy. 
  - means not available. The best results are in bold. 
  $^*$ represents that the result is reported in the corresponding reference.}
  \label{tab:result}
  \begin{tabular}{|l|p{2.5cm}<{\centering}|p{3.5cm}<{\centering}|p{2cm}<{\centering}|}
  \hline
      \multicolumn{1}{|c|}{\textbf{Models}}     & \textbf{Toys \& Games}    & \textbf{Sports \& Outdoors}       &  \textbf{News} \\
  \hline
  TextCNN \cite{kim-2014-convolutional}           & 70.5        & 72.0  & 77.5     \\  
  Bi-LSTM \cite{tang-etal-2016-effective}         & 70.7\        & 71.9    & 76.5  \\
   BERT \cite{devlin-etal-2019-bert}              & 75.5{ }     & 74.2 & 85.7   \\
   BERT(head+tail)\cite{DBLP:conf/cncl/SunQXH19}  & 75.9{ }     & 74.0   & 86.1    \\
  %  \hline
   HAN  \cite{yang2016hierarchical}                      &   69.1{ }  & 72.3 & 78.6  \\ 
   CAHAN \cite{DBLP:journals/corr/abs-1908-06006}   & 70.8{ }   & 73.0  & 79.9        \\
  \hline
  % \multirow{3}{*}{Summary-based} &
   HSSC   \cite{DBLP:conf/ijcai/MaSLR18}          & 71.9$^*$     & 73.2$^*$  & -      \\ 
   SAHSSC  \cite{pmlr-v95-wang18b}                & 72.5$^*$   &73.6$^*$   & -    \\ 
  \hline
  % \multirow{2}{*}{Proposed model}& 
  
  \textbf{HIN}                                 & \textbf{77.5}           & \textbf{76.7} &  \textbf{89.0}     \\
  \textbf{HIN-SR }                             & \textbf{78.1}    &  \textbf{77.2}  & \textbf{89.3}  \\
  
  \hline
  \end{tabular}
  \end{table}

\subsection{Ablation Study of HIN}
For further study, we conduct a series of ablation experiments to
examine the relative contributions of each component in HIN on accuracy ({Acc}) and macro-F1 ({F1}).
To this end, HIN is compared with the following variants:
\begin{itemize}
    \item \textbf{w/o Doc.} indicates the HIN model without modeling the document-level interactions
    via removing attention mechanism.
    \item \textbf{w/o Doc. and Seg.} indicates the HIN model removing Bi-GRU and without modeling both segment-level and document-level interactions. 
    \item \textbf{w/o Interactions } indicates the model
    performs the sentiment classification by directly concatenating the summary and document embeddings.  
    \item \textbf{w/o Summary} indicates the model performs the sentiment classification by BERT without the summary and corresponding interactions. 
\end{itemize}

The results are shown in Table \ref{tab:abaltionstudy}. 
We observe that compared with these partial models,
the full model yields significant improvements on F1. 
Noted that HIN without both document- and segment-level interactions
achieve the best accuracy in Toys \& Games dataset,
a reasonable explanation is due to the imbalanced distribution of categories.
Therefore, consideration the imbalanced distribution, F1 is a comprehensive method for evaluation.

\begin{table}
  \caption{An ablation study of the proposed model. 
  Evaluation metrics are accuracy (Acc) and F1. 
  The best results are in bold.
  }
  \label{tab:abaltionstudy}
  \centering
  % \resizebox{\linewidth}{!}{$
  \begin{tabular}{
      |l|
      p{1.4cm}<{\centering}|
      p{1.4cm}<{\centering}|
      p{1.6cm}<{\centering}|
      p{1.6cm}<{\centering}|
      p{1.2cm}<{\centering}|
      p{1.2cm}<{\centering}|}
  \hline
  \multirow{2}{*}{\textbf{Ablation Settings}} 
  & \multicolumn{2}{c|}{\textbf{Toys \& Games}} & \multicolumn{2}{c|}{\textbf{Sports \& Outdoors} } &  \multicolumn{2}{c|}{\textbf{News} }\\
  \cline{2-7}
  & {Acc} & {F1}   & { Acc }  & { F1}  & { Acc }  &{ F1}    \\      
  \hline
  $\textbf{HIN}$        &{77.5}          & \textbf{54.8} & \textbf{76.7}     & \textbf{62.7}   & \textbf{89.0}  & \textbf{81.3}      \\
   - w/o Doc.           & 76.9           & 53.0          & 76.2              & 62.3            & 88.9           &  81.1            \\     
   - w/o Doc. and Seg.  & \textbf{77.8}  & 40.0          & 75.9              & 62.2            & 87.9           &  80.2          \\     
   - w/o Interactions   & 76.0           & 50.6          & 75.3              & 61.7            & 87.8           &  79.6         \\    
   - w/o Summary        & 75.5           & 49.3          & 74.2              & 61.0            & 85.7           &  78.1       \\        
 \hline 
  \end{tabular}
  % $}
  
  \end{table}

% ZONG
HIN without document-level interactions obtained a worse result than the complete model.
The fact can be attributed that the module can aggregate multiple segments with rich subject information with the consideration of the summary representation via the attention mechanism.
% HIN
Moreover, HIN without both document- and segment-level interactions reports a relatively poor performance as Bi-GRU was able to capture more contextual information to generate better summary and segment representations via the propagation in the hidden states.
% no
Without modeling any interactions, the performance of HIN drops off significantly, which demonstrates the effectiveness of interactions between the summary and document and the encoding can indeed capture more effective context semantic features with the utilization of a user-generated summary.
Besides, for HIN without any interactions, modeling with the summary outperforms modeling without the summary, which reveals that the summary can effectively serve the identification of sentiment polarity.

\section{Discussion} \label{t:analysis}
In this section, we give some analyses about training episodes and document length on predicting sentiment labels, and discuss the effect of the summary with visualization.

\subsection{Analysis of Training Episodes}

\begin{figure}[h]
    \centering
    \includegraphics[width=0.9\linewidth]{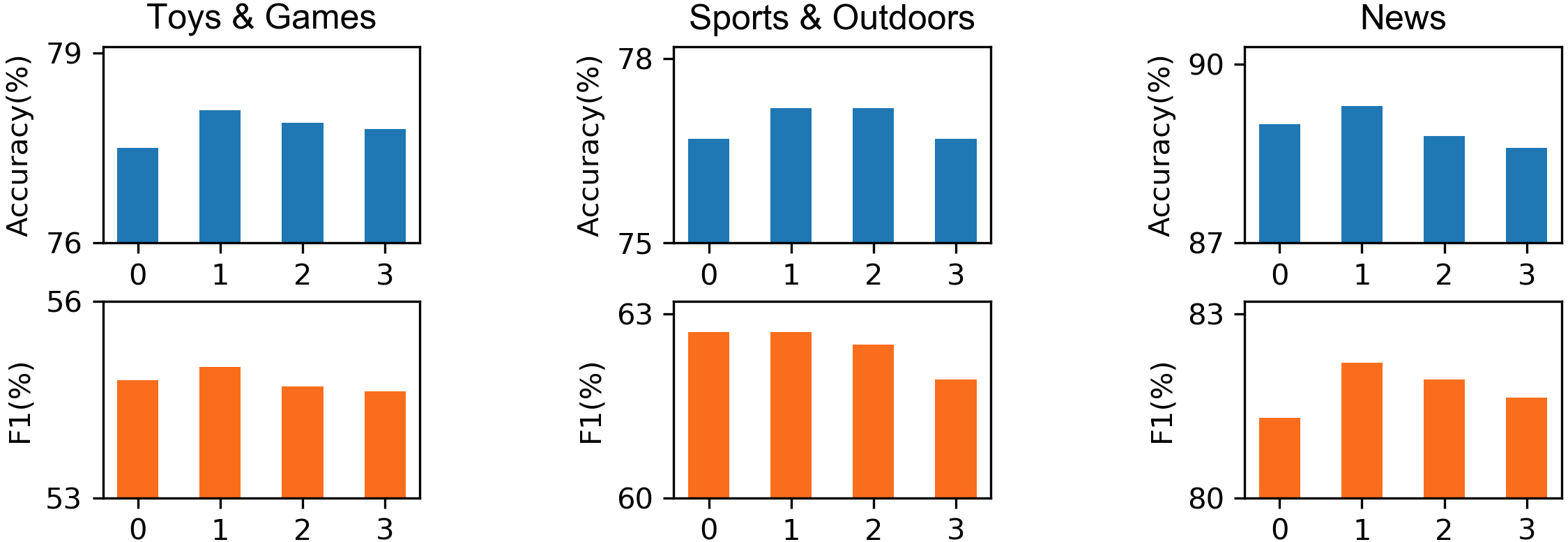}
    \caption{Results against training episodes. The X-axis is the number of training episodes and the Y-axis represents accuracy or F1. }
    \label{fig:iteration}
\end{figure}

The number of episodes is an important parameter in the sentiment-based rethinking mechanism (SR). 
We investigate its effects on three datasets in Fig.~\ref{fig:iteration}.

From the results in Fig.~\ref{fig:iteration}, we observe that
as the number of training episodes increases, the performance grows significantly and then declines slightly on both accuracy and F1. And the best performance is when the number of training episodes is set to 1.

Superior performance when training episode is set to 1 than 0 
demonstrates the effectiveness of gold sentiment label information. 
Rethinking these high-level sentiment patterns guides the model to capture more discriminative features specifically for different classes.
However, 
drop performance when the number of training episodes increases reveals that, 
an overload of sentiment label information would 
limit the original feature expression of HIN 
and result in a slightly descending performance. 
Besides, three datasets show different meliorations of performance when the number of training episodes is set to 1 than 0,
Especially, the model achieves improvements of 0.9\% on in News dataset where more vague semantic links in Chinese bring much noise samples.
This reveals that the SR can release the negative impacts of noisy samples simultaneously.

\subsection{Analysis of Document Length} 
\begin{figure}[h]
    \centering
    \subfigure[\text{Toys \& Games} dataset.]{
    \includegraphics[width=5.5cm]{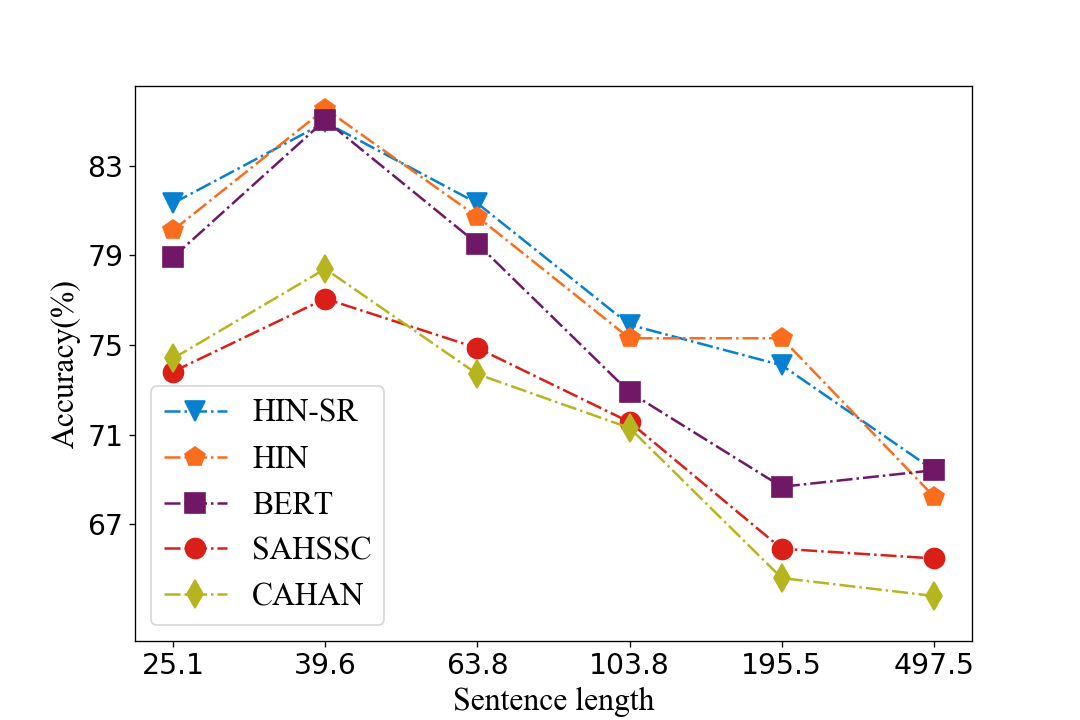}
    }
    % \quad
    \subfigure[\text{Sports \& Outdoors} dataset.]{
    \includegraphics[width=5.5cm]{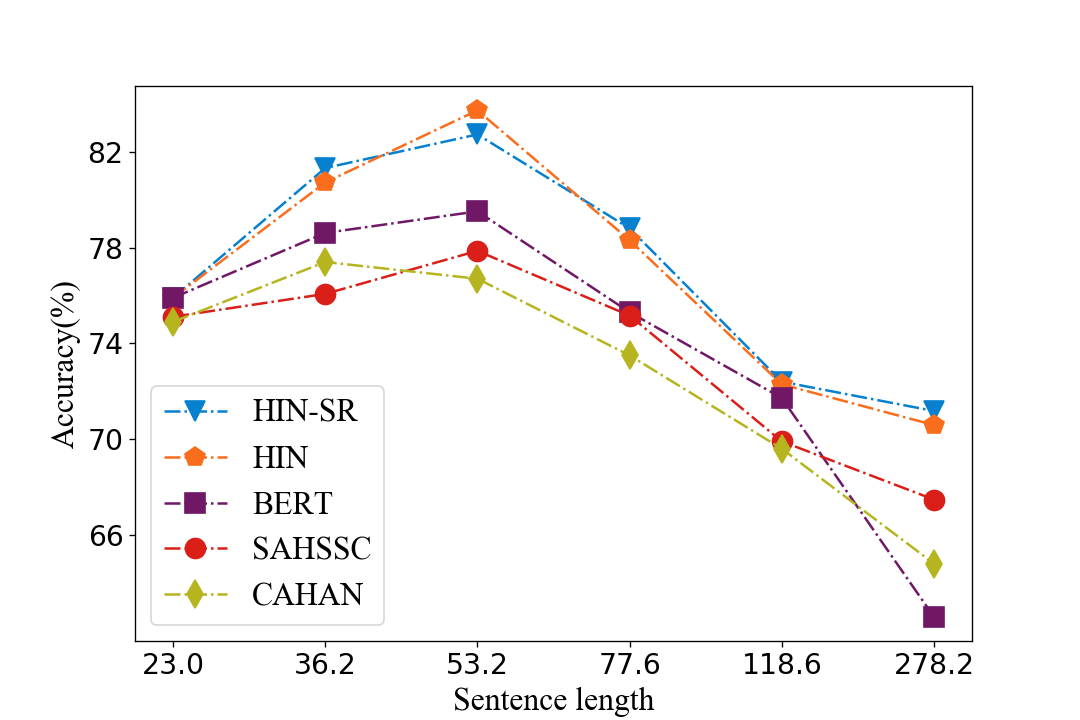}
    }
    % \quad
    \subfigure[News dataset.]{
    \includegraphics[width=5.5cm]{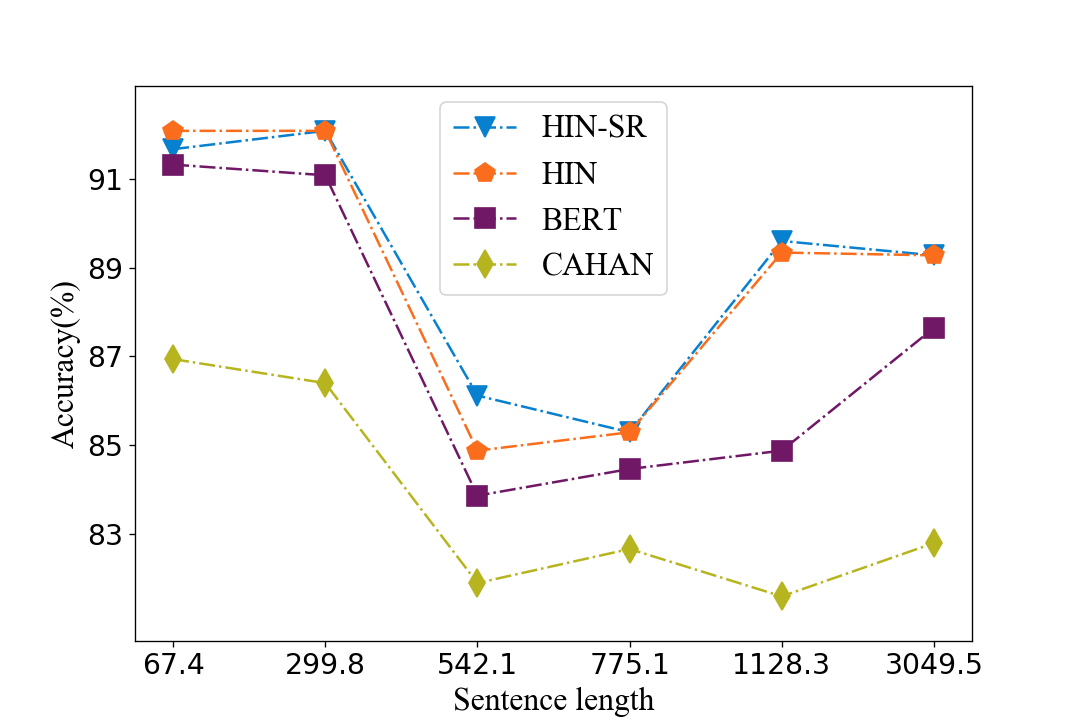}
    %\caption{fig1}
    }
    \caption{ Results against document length. We divide the testing dataset into six parts according to the quantiles of document length.
    The X-axis is the average length of each part and the Y-axis indicates accuracy. }
    \label{length}
\end{figure}

We further investigate the performance of HIN-SR, HIN, and several baselines 
when analyzing sentiment polarity with different document lengths.
The results of different datasets are shown in 
Fig.~\ref{length}.

As document length increases, the performance of all models clearly decreases 
after a certain range. 
The result reveals that document-level sentiment analysis is more challenging to 
the longer document usually containing more vague semantic links and complicated sentiment information.
In particular, the curve of accuracy presents a partial upward trend 
over an extremely long length in the News datasets.
And we speculate that more semantic information can partially alleviate 
the above phenomenon.

From the figures, it is seen that HIN achieves a considerable improvement than other baselines on three datasets over almost any range of document length. 
Compared with other baselines system, HIN takes 
the multiple-granularity interactions between the summary and document into considerations.
After modeling the interactions, HIN can adequately exploit the semantic 
and subject information of the summary and thus learn subject-oriented document representation 
for document-level sentiment analysis. 
In addition, in Fig.~\ref{length}(a) and (c), 
the results of BERT are slightly higher than HIN at a few of length ranges.
Through the analysis of the abnormal samples, we conclude 
that this is probably due to the uneven distribution of samples.

Moreover, HIN-SR slightly outperforms than HIN over most length ranges. 
This indicates that 
the rethinking mechanism introducing sentiment label information can 
adapt and refine the HIN with high-level sentiment features to boost the 
performance.
In addition, HIN-SR is not sensitive to document length and is adaptive in both short and long documents.  
It illustrates the robustness of the model becomes better with the help of reweighting document representations.
We contribute it to this rethinking mechanism can indeed alleviate the data noise
via feedbacking the sentiment label information.

\begin{figure}[t]
  \centering
  \includegraphics[width=0.9\linewidth]{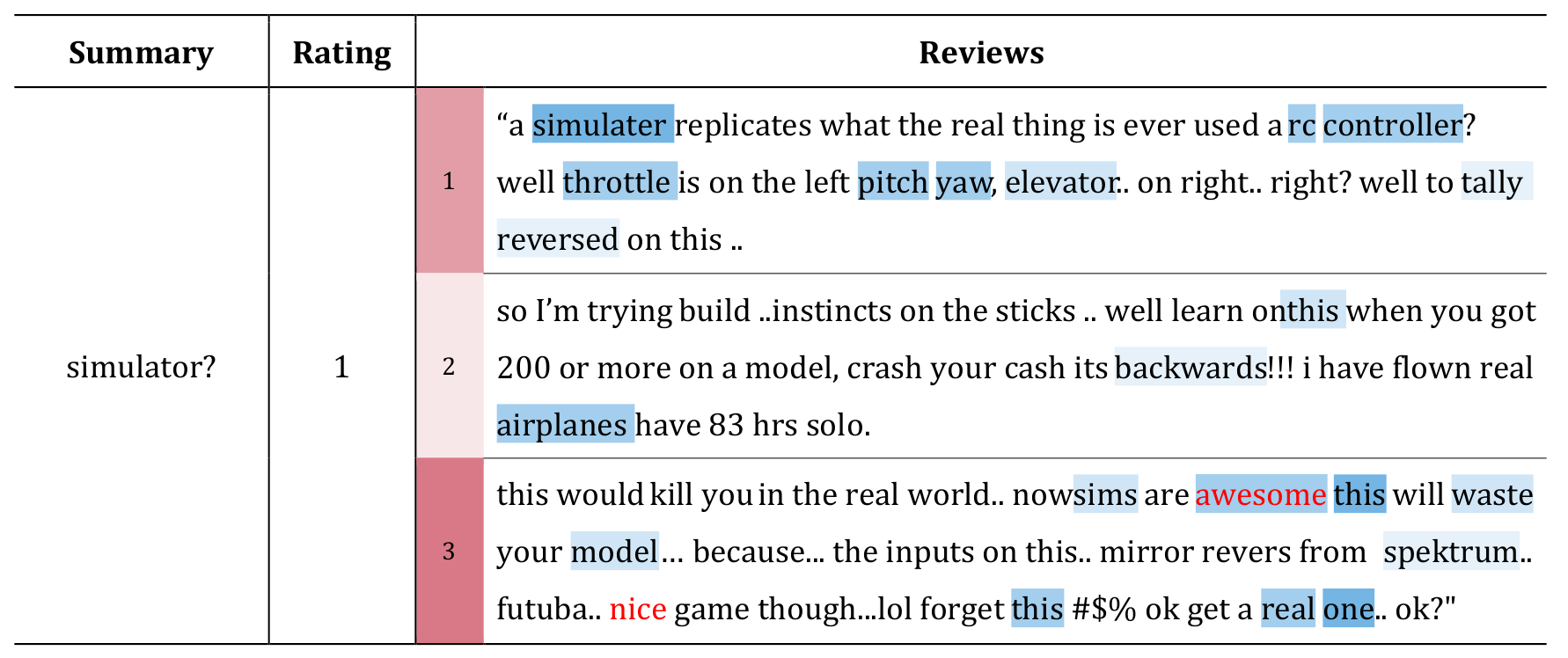} %case2
  \caption{ 
  The review is from Toys \& Games dataset with a negative sentiment label. 
  Each block represents one segment candidate of the document.
  Blue color denotes the word weight after character- and segment-level interaction encoding.
  Red color indicates candidate weight in the document-level interaction encoding module.
  }\label{reviews}
\end{figure}

\subsection{Case Study}

We visualize multi-granularity interactions encoding modules in Fig.~\ref{reviews}.
The review describes the disappointing experience of using a simulator,
containing complex sentiment polarity expressed such as {\it awesome} and {\it nice}.
According to the summary, the major point of the review can conclude as {\it a simulator}.

From the blue marks, we can observe that after encoding character- and segment-level interactions between the summary and document,
our model performs by consciously selecting 
texts carrying the major point (i.e., {\it simulator}).
Note that the contexts of these subject-related tokens contain pivotal sentimental information, such as {\it awesome}. While the sentiment expressed {\it nice} is neglected because 
it is irrelevant to the subject {\it simulator}.
Subsequently, document-level interaction features are captured 
to further aggregate the document representation for sentiment classification.

In conclusion, HIN successfully tackles the challenge of
vague semantic links and complicate sentiment information
and makes correct classification. 
Evidenced by the visualization, subject information
is affirmative for capturing accurate sentiment information. 
Our proposed model can explicitly explore multi-granularities interactions between the summary and document to learn a subject-oriented document representation.

\section{Conclusion} \label{t:con}
In this work, we have investigated the task of document-level sentiment analysis. 
We design a hierarchical interaction networks (HIN) model to learn a subject-oriented document representation for sentiment classification.
The main idea of HIN is 
to exploit bidirectional interactions between 
the user-generated summary and document. 
Furthermore, a sentiment-based rethinking mechanism (SR) refines the weights of document features to learn a more sentiment-aware document representation and alleviate the negative impact of noisy data.
Experimental Results on three widely public datasets have demonstrated that HIN-SR outperforms significantly and tackles long documents with the vague semantic links and abundant sentiments effectively.
The proposed model is of great significance to DSA and related applications.

For future work, we will consider more interaction information, (e.g., interactions between documents) and more detailed theoretical analysis of rethinking mechanism, to further improve the performance.

% \bibliography{refer}
%

\end{document}